\documentclass[final]{cvpr}

\usepackage{times}
\usepackage{epsfig}
\usepackage{graphicx}
\usepackage{amsmath}
\usepackage{amssymb}

\usepackage{multirow}
\usepackage{multicol}
\usepackage{makecell}
\usepackage{fancyhdr}

\newcommand{\ourmodel}{BSUV-Net 2.0}
\newcommand{\ourmodelfast}{Fast BSUV-Net 2.0}
\usepackage{pifont}
\newcommand{\cmark}{\ding{51}}%
\newcommand{\emptysub}{E}
\newcommand{\recentsub}{R}
\newcommand{\currentsub}{C}
\newcommand{\fgsub}{{FG}}
\newcommand{\anysub}{k}

\newcommand{\emptyref}{\mathbf{I_\emptysub}}
\newcommand{\recentref}{\mathbf{I_\recentsub}}
\newcommand{\currentfr}{\mathbf{I_\currentsub}}
\newcommand{\fg}{\mathbf{I_\fgsub}}
\newcommand{\anyfr}{\mathbf{I}_\anysub}

\newcommand{\emptyrefcrop}{\mathbf{\widetilde{I}_\emptysub}}
\newcommand{\recentrefcrop}{\mathbf{\widetilde{I}_\recentsub}}
\newcommand{\currentfrcrop}{\mathbf{\widetilde{I}_\currentsub}}
\newcommand{\fgcrop}{\mathbf{\widetilde{I}_\fgsub}}
\newcommand{\anyfrcrop}{{\mathbf{\widetilde{I}}}_\anysub}

\newcommand{\iomsuper}{IO}
\newcommand{\emptyrefiom}{\mathbf{\widetilde{I}_\emptysub^{\iomsuper}}}
\newcommand{\recentrefiom}{\mathbf{\widetilde{I}_\recentsub^{\iomsuper}}}
\newcommand{\currentfriom}{\mathbf{\widetilde{I}_\currentsub^{\iomsuper}}}
\newcommand{\fgiom}{\mathbf{\widetilde{I}_\fgsub^{\iomsuper}}}
\newcommand{\anyfriom}{\mathbf{\widetilde{I}^{\iomsuper}}_\anysub}

\newcommand{\emptyreffinal}{\mathbf{\widehat{I}_\emptysub}}
\newcommand{\recentreffinal}{\mathbf{\widehat{I}_\recentsub}}
\newcommand{\currentfrfinal}{\mathbf{\widehat{I}_\currentsub}}
\newcommand{\fgfinal}{\mathbf{\widehat{I}_\fgsub}}
\newcommand{\anyfrfinal}{\mathbf{\widehat{I}}_\anysub}

\newcommand{\width}{w}
\newcommand{\height}{h}

\newcommand{\widthcrop}{\widetilde{\width}}
\newcommand{\heightcrop}{\widetilde{\height}}

\newcommand{\cropfunc}{\mathcal{C}}
\newcommand{\resizefunc}{\mathcal{R}}

\usepackage[pagebackref=true,breaklinks=true,colorlinks,bookmarks=false]{hyperref}

\begin{document}

%%%%%%%%% TITLE
\title{BSUV-Net 2.0: Spatio-Temporal Data Augmentations for Video-Agnostic Supervised Background Subtraction}

\author{M. Ozan Tezcan, Prakash Ishwar, Janusz Konrad
\thanks{This work was supported in part by ARPA-E under agreement DE-AR0000944 and by the donation of Titan GPUs from NVIDIA Corp.}
\thanks{This work has been submitted to the IEEE for possible publication. Copyright may be transferred without notice, after which this version may no longer be accessible.}
\\
Boston University, Boston, MA 02215\\
{\tt\small \{mtezcan, pi, jkonrad\}@bu.edu}
}

\maketitle

\begin{abstract}
Background subtraction (BGS) is a fundamental video processing task which is a key component of many applications. Deep learning-based supervised algorithms achieve very good perforamnce in BGS, however, most of these algorithms are optimized for either a specific video or a group of videos, and their performance decreases dramatically when applied to unseen videos. Recently, several papers addressed this problem and proposed video-agnostic supervised BGS algorithms. However, nearly all of the data augmentations used in these algorithms are limited to the spatial domain and do not account for temporal variations that naturally occur in video data. In this work, we introduce spatio-temporal data augmentations and apply them to one of the leading video-agnostic BGS algorithms, BSUV-Net. We also introduce a new cross-validation training and evaluation strategy for the CDNet-2014 dataset that makes it possible to fairly and easily compare the performance of various video-agnostic supervised BGS algorithms. Our new model trained using the proposed data augmentations, named BSUV-Net 2.0, significantly outperforms state-of-the-art algorithms evaluated on unseen videos of CDNet-2014. We also evaluate the cross-dataset generalization capacity  of BSUV-Net 2.0 by training it solely on CDNet-2014 videos and evaluating its performance on LASIESTA dataset. Overall, BSUV-Net 2.0 provides a $\sim$5\% improvement in the F-score over state-of-the-art methods on unseen videos of CDNet-2014 and LASIESTA datasets. Furthermore, we develop a real-time variant of our model, that we call Fast BSUV-Net 2.0, whose performance is close to the state of the art. 
%The source code of BSUV-Net 2.0 will be made publicly available should the paper be accepted for publication.
\end{abstract}

% \begin{keywords}
% Background subtraction, foreground detection, scene independent, scene agnostic, deep learning, data augmentation
% \end{keywords}

% \titlepgskip=-15pt

\maketitle

\section{Introduction}
\label{sec:intro}

Background subtraction (BGS) is one of fundamental video processing blocks frequently used in applications such as advanced video surveillance, human activity recognition, {autonomous navigation}, etc. \cite{bouwmans2019deep, garcia2020bgsappsurvey}. BGS can be defined as binary video segmentation aiming to extract foreground and background regions in each video frame. 

End-to-end BGS algorithms can be loosely grouped into three categories: (i) Unsupervised algorithms, (ii) \textit{video-} or \textit{video-group-optimized} supervised algorithms and (iii) \textit{video-agnostic} supervised algorithms. Unsupervised algorithms attempt to mathematically model the background and extract the foreground pixels accordingly. Several model-based approaches in this category, such as PAWCS \cite{st2015pawcs} and WiseNetMD \cite{lee2018wisenetmd}, achieve very competitive performance, however they are currently outperformed by deep learning-based supervised algorithms.

Several recent papers introduced {\it video-} or {\it video-group-optimized} algorithms \cite{zeng2018mfcnnbgs, bakkay2018bscgan, lim2018fgsegnetv3, babaee2018cnnbgs, sakkos20173dbgs} and {\it video-agnostic} \cite{tezcan2020bsuv,mandal20193dfr} supervised BGS algorithms. The first category report results for methods that have been trained and tested on the same set of videos and their performance on unseen videos is not reported. 
On the other hand, {\it video-agnostic} algorithms report results on unseen videos by training and testing on disjoint sets of videos. One of the most successful {\it video-agnostic} BGS algorithms, BSUV-Net, uses spatial and semantic information from different time scales to improve performance on unseen videos. However, due to limited amount of labeled BGS data, BSUV-Net's performance on very challenging scenarios is still insufficient for real-world applications. 

One of the most successful approaches for increasing the generalization capacity of computer vision algorithms trained with limited data is the use of data augmentation. Spatial data augmentation, such as random crops, rotations, color changes, etc.\ have proved very successful in image-related tasks \cite{taylor2017improving, shorten2019survey}. A simple \textit{spatio-temporal} data augmentation was introduced in BSUV-Net \cite{tezcan2020bsuv} to handle illumination differences between videos and resulted in some performance improvement. However, to the best of our knowledge, besides that work there are no other data augmentation attempts tailored to BGS that make use of both spatial and temporal information in a comprehensive manner. In this paper, we introduce a comprehensive suite of spatio-temporal data augmentation methods and adapt them to BSUV-Net. The proposed augmentations address some key BGS challenges, such as PTZ (pan-tilt-zoom) operation, camera jitter and presence of intermittently-static objects. We conduct {\it video-agnostic} performance analysis and show that these data augmentations significantly increase algorithm's performance for targeted categories without any significant loss of performance in other categories. We also show that a network trained on a combination of several spatio-temporal data augmentations outperforms state of the art (SOTA) methods on unseen videos by $\sim$5\% in terms of F-score. 
%
%\piedit{please double-check that this is really a 5 percent improvement and not a 5 percentage-point improvement.}
%\ozan{It is 0.8387 vs 0.7986 (Table 5), so I guess it should be 5 percent improvement by (0.8387 - 0.7986) / 0.7986 $\sim\!\!$ 0.05}
%
Furthermore, we demonstrate a cross-dataset generalization capacity of \ourmodel{} by training it solely on CDNet-2014 and testing it on LASIESTA \cite{cuevas2016lasiesta}, a completely unseen dataset.
Here too, \ourmodel{} outperforms SOTA methods by $\sim$5\% (F-score). The main contributions of this work are as follows:
% \vspace{-1ex}
\begin{enumerate} %\topsep -5pt\partopsep -5pt\itemsep -4pt
    \item {\bf Spatio-Temporal Data Augmentation: } We introduce spatio-temporal data augmentation methods for BSUV-Net to mimic challenging BGS scenarios, such as PTZ operation, camera jitter and presence of intermittently-static objects (e.g., cars stopped at a streetlight). Our experimental results show that these augmentations significantly improve the performance on unseen videos of corresponding categories. 
    
    \item {\bf Fair evaluation strategy for CDNet-2014: } Although CDNet-2014 is an extensive BGS dataset, it lacks a training/testing split for use by supervised learning approaches. We introduce a split of CDNet-2014 videos into 4 groups to be used for cross-validation. In this way, we can easily evaluate any supervised BGS algorithm on all CDNet-2014 videos in a video-agnostic manner. This will simplify algorithm performance comparisons in the future.
    
    \item {\bf State-of-the-art and real-time results: } Our proposed algorithm outperforms SOTA on CDNet-2014 and LASIESTA for unseen videos by $\sim \!\! 5 \%$. We also introduce a real-time variant of \ourmodel{} which runs at $\sim \! 29$ FPS and performs on-par with SOTA. We will publicly share our training and validation scripts as well as the trained models should the paper be accepted for publication.
\end{enumerate}

%-------------------------------------------------------------------------
\section{Related Work}
\label{sec:related}

{\bf Unsupervised BGS algorithms:} The early attempts at BGS have relied on probabilistic background models such as Gaussian Mixture Model (GMM) \cite{stauffer1999gmm} and Kernel Density Estimation (KDE) \cite{elgammal2002kde}. Following the idea of BGS based on background modeling, more sophisticated algorithms were introduced (e.g., SubSENSE \cite{st2015subsense}, PAWCS \cite{st2015pawcs}, SWCD\cite{icsik2018swcd} and WisenetMD). 
Recently, VTD-FastICA \cite{huang2020icabgs} has applied independent component analysis to multiple frames, whereas Giraldo and Bouwmans have introduced a graph-based algorithm that considers the instances in a video as nodes of a graph and computes BGS predictions by minimizing the total variation \cite{giraldo2020graphbgs, giraldo2020semi}.
Finally, RT-SBS \cite{cioppa2020rtsbs} and RTSS \cite{zeng2019rtss} have combined unsupervised BGS algorithms with deep learning-based semantic segmentation algorithms, such as PSPNet \cite{zhao2017pspnet}, to improve BGS predictions.

{\bf {Video-} or {video-group-optimized} supervised BGS algorithms:} The early attempts of deep learning at BGS have focused on \textit{video-} or {\it video-group-optimized} algorithms, which are tested on the same videos that they are trained on. Usually, their performance on unseen videos is not reported. \textit{Video-optimized} algorithms train a new set of weights for each video using some of the labeled frames from this very test video, while \textit{video-group-optimized} ones train a single network for the whole dataset by using some labeled frames from the whole dataset. They all achieve near-perfect results \cite{bakkay2018bscgan, lim2018fgsegnetv3, babaee2018cnnbgs, patil2020end}. Although these algorithms might be very useful for speeding up the labeling process of new videos, their performance drops significantly when they are applied to unseen videos \cite{tezcan2020bsuv, mandal20203dcd, mandal20193dfr}. Clearly, they are not suitable for real-world applications.

{\bf Video-agnostic supervised BGS algorithms:} Recently, several supervised-BGS algorithms for unseen videos have been introduced. ChangeDet \cite{mandal2020scene}, 3DFR \cite{mandal20193dfr}, and 3DCD \cite{mandal20203dcd} proposed end-to-end convolutional neural networks for BGS that use both spatial and temporal information based on previous frames and a simple median-based background model. Similarly, Kim {\it et al.}\ \cite{kim2020foreground} introduced a U-Net-based \cite{ronneberger2015unet} neural network that uses a concatenation of the current frame and several background models generated at different time scales as the input.
During evaluation, these methods divided videos of a popular BGS dataset, CDNet-2014 \cite{goyette2012changedetection}, into a training set and a testing set, and reported results for the test videos, unseen by the algorithm during training. Although all three algorithms outperform unsupervised algorithms on their own test sets, their true performance is unknown since no results were reported for the full dataset. Furthermore, these algorithms cannot be compared with each other since each used a different train/test split. Recently,
BSUV-Net \cite{tezcan2020bsuv} also used U-Net architecture but added a novel augmentation step to handle illumination variations, thus improving performance on unseen videos. The authors proposed an evaluation approach on the whole CDNet-2014 dataset by using $18$ training/test sets and demonstrated one of the top performances on the dataset's evaluation server. However, their approach with 18 training/test sets is complicated and makes any comparison of future algorithms against BSUV-Net difficult. 

{\bf BGS versus Video-Object Segmentation:} Several video-object segmentation datasets and challenges have been introduced in the last few years \cite{xu2018youtubevos, yang2019youtubevis, perazzi2016davis, pont2017davissemi, caelles2019davisunsupervised}. That body of work can be easily confused with BGS. However, there exist important differences that make video-object segmentation algorithms not suitable for BGS. For example, the unsupervised multi-object segmentation dataset in the DAVIS Challenge \cite{caelles2019davisunsupervised} includes annotations only for objects that capture human attention the most. No annotations are provided for foreground objects that are not in the focus of attention. Since one of the main applications of BGS is video surveillance, the detection of every moving object is critical (e.g., a person on whom the video is not focused may still act suspiciously).
Furthermore, the semi-supervised multi-object segmentation dataset in the DAVIS Challenge assumes that semantic segmentation of the first frame of each video is available. Youtube-video object/instance segmentation datasets \cite{xu2018youtubevos, yang2019youtubevis} include annotations for objects from a set of predefined categories. The annotations for a single category might include both stationary background objects and moving foreground objects (e.g., there is no distinction between a parked motorbike [background object] and a moving motorbike [foreground object]). Thus, a comparison of object/instance segmentation and BGS is not very meaningful. In this paper, we do not investigate algorithms designed for video-object segmentation since, despite similarities, they focus on a different task.
%\jlk{More broadly, do video object/instance segmentation methods predict segmentation masks or bounding boxes? Those that find bounding boxes only are not comparable to BGS. Those that find segmentations may be argued against as discussed above}\ozan{All the datasets that I mentioned in this paragraph have segmentation masks. I didn't include bounding box datasets at all since I don't think they can be confused with BGS.}

In this paper, we improve the performance of BSUV-Net by introducing multiple spatio-temporal data augmentations designed to attack the most common challenges in BGS. We name our improved algorithm \ourmodel{} and show that it significantly outperforms state-of-the-art BGS algorithms on unseen videos. We also introduce a real-time version of \ourmodel{} and call it \ourmodelfast{}. Finally, we propose a 4-fold cross-validation strategy to facilitate fair and streamlined comparison of unsupervised and {\it video-agnostic} supervised algorithms, which should prove useful for future BGS algorithm comparisons on CDNet-2014.

%-------------------------------------------------------------------------
\section{Summary of BSUV-Net}

BSUV-Net is among the top-performing BGS algorithms designed for unseen videos. We briefly summarize it below.

BSUV-Net is a U-Net-based \cite{ronneberger2015unet} CNN which takes a concatenation of 3 images as input and produces a probabilistic foreground estimate. The input consists of two background models captured at different time scales and the current frame. One background model, called ``empty'', is a manually-selected static frame void of moving objects, whereas the other model, called ``recent'', is the median of previous 100 frames. All three input images consist of 4 channels: R, G, B color channels and a foreground probability map (FPM). FPM is an initial foreground estimate for each input image computed by DeepLabv3 \cite{chen2017deeplab}, a semantic segmentation algorithm that does not use any temporal information. For more details on network architecture and the FPM channel of BSUV-Net, please refer to the original paper \cite{tezcan2020bsuv}. BSUV-Net uses relaxed Jaccard index as the loss function: 
{\small
\begin{equation*}
J_R(\mathbf{Y}, \widehat{\mathbf{Y}}) \!=\!
\frac{T + \sum\limits_{m,n}(\mathbf{Y}[m,n] \widehat{\mathbf{Y}}[m,n])}{T \!+\!  \sum\limits_{m,n}\!\!\Big(\mathbf{Y}[m,n] \!+\! \widehat{\mathbf{Y}}[{m,n}] \!-\! \mathbf{Y}[{m,n}]  \widehat{\mathbf{Y}}[{m,n}] \Big)}
\end{equation*}}%
where $\widehat{\mathbf{Y}} \in [0, 1]^{w \times h}$ is the predicted foreground probability map, $\mathbf{Y} \in \{0, 1\}^{w \times h}$ is the ground-truth foreground label, $T$ is a smoothing parameter and $m$, $n$ are spatial locations.

The authors of BSUV-Net also proposed a novel data-augmentation method for video that addresses illumination differences (ID) often present between video frames. In an ablation study, they demonstrated a significant impact of this augmentation on the overall performance.
In this paper, we expand on this idea and introduce a new category of data augmentations designed specifically for spatio-temporal video data. 

%-------------------------------------------------------------------------
\section{Spatio-Temporal Data Augmentations}
\label{sec:augmentations}

In this section, we first introduce mathematical notation. Then, we propose new spatio-temporal augmentations and also describe the illumination-difference augmentation which was proposed in BSUV-Net. Fig.~\ref{fig:aug} shows one example of each of the proposed augmentations.
%-----------------------------------------------------------------------
\subsection{Notation}
\label{sec:formulation}

Let us consider an input-label pair of BSUV-Net. The input consists of $\emptyref, \recentref, \currentfr \in \mathbb{R}^{\width \times \height \times 4}$ an empty background, a recent background and the current frame, respectively, where $\width, \height$ are the width and height of each image. Each image has 4 channels: three colors (R, G, B) plus FPM discussed above. Similarly, let $\fg \in \{0, 1\}^{\width \times  \height}$ be the corresponding foreground label field where $0$ represents the background and $1$ -- the foreground. 

Although the resolution of input images varies from video to video, it is beneficial to use a single resolution during training in order to leverage parallel processing of GPUs. Therefore, the first augmentation step we propose is {\it spatio-temporal cropping} that maps each video to the same spatial resolution. In the second step, we propose two additional augmentations that modify video content but not size.

In our two-step process, in the first step we use different cropping functions to compute $\emptyrefcrop, \recentrefcrop, \currentfrcrop \! \in \! \mathbb{R}^{\widthcrop \times \heightcrop \times 4}$ and {$\fgcrop \!\! \in \!\! \{0, 1\}^{\widthcrop \times \heightcrop}$} from $\emptyref, \recentref, \currentfr$ and $\fg$ where $\widthcrop, \heightcrop$ are the desired width and height after cropping. In the second step, we apply post-crop augmentations to compute $\emptyreffinal, \recentreffinal, \currentfrfinal \! \in \! \mathbb{R}^{\widthcrop \times \heightcrop \times 4}$ and $\fgfinal \!\! \in \!\! \{0, 1\}^{\widthcrop \times \heightcrop}$ from $\emptyrefcrop, \recentrefcrop, \currentfrcrop$ and $\fgcrop$. Below, we explain these two steps in detail.

\subsection{Spatio-Temporal Crop}
\label{sec:randomcrop}

Here we describe 3 augmentation techniques to compute $\emptyrefcrop, \recentrefcrop, \currentfrcrop, \fgcrop$ from $\emptyref, \recentref, \currentfr, \fg$, each addressing a different BGS challenge. We begin by defining a cropping function, to be used in this section, as follows:
%\vspace{-1 ex}
{\small
\begin{equation*}
% \label{eqn:alignedcrop}
    \cropfunc(\mathbf{I}, i, j, h, w) = 
     \mathbf{I}\left[\left\lceil i-\tfrac{h}{2} \right\rceil : \left\lceil  i+\tfrac{h}{2}\right\rceil,  \left\lceil  j-\tfrac{w}{2} \right\rceil : \left\lceil  j+\tfrac{w}{2}\right\rceil ,  1:4\right]
 \end{equation*}
}%
\noindent where $i, j$ are the center coordinates, $h, w$ are height and width of the crop, $\lceil \cdot \rceil$ denotes the ceiling function and $a:b$ denotes the range of integer indices $a,a+1,\ldots,b-1$.
% \jlk{So it includes $a$ but not $b$?}
% \ozan{Yes.}

\textbf{Spatially-Aligned Crop:}
This is an extension of the widely-used spatial cropping for individual images. Although this is {straightforward, we} provide a precise definition in order to clearly define steps in subsequent sections.

The output of a spatially-aligned crop is defined follows:
%
%\vspace{-1 ex}
{\small
\begin{equation*}
% \label{eqn:alignedcrop}
     \anyfrcrop = \cropfunc(\anyfr,\ i,\ j,\ \heightcrop,\ \widthcrop)
     \quad \text{ for all }
     k \in \{\mathbf{\emptysub},\ \mathbf{\recentsub},\ \mathbf{\currentsub},\ \mathbf{\fgsub}\},
 \end{equation*}
}%
\noindent where $i, j$ are randomly-selected spatial indices of the center of the crop. This formulation allows us to obtain a fixed-size, spatially-aligned crop from the input-label pair. 

\textbf{Randomly-Shifted Crop:}
One of the most challenging scenarios for BGS algorithms is camera jitter which results in random spatial shifts between consecutive video frames. However, since the variety of such videos is limited in public datasets, it is not trivial to learn the behavior of camera jitter using a data-driven algorithm. In order to address this, we introduce a new data augmentation method by simulating camera jitter. As a result, spatially-aligned inputs become randomly shifted. This is formulated as follows:
%
%\vspace{-1 ex}
{\small
\begin{equation*}
    \anyfrcrop = \cropfunc(\anyfr,\ i_k,\ j_k,\ \heightcrop,\ \widthcrop)
    \quad \text{ for all }
     k \in \{\mathbf{\emptysub},\ \mathbf{\recentsub},\ \mathbf{\currentsub},\ \mathbf{\fgsub}\},
\end{equation*}
%\vglue -1 ex
}%
\noindent where $i_k, j_k$ are randomly-selected, but such that $i_\currentsub = i_\fgsub$ and $j_\currentsub = j_\fgsub$
%\jlk{This is a bit sloppy notation since subscript is a frame type. Perhaps $i_k^C$?}\ozan{I am not sure what you mean by $k$ in $i_k^C$. I tried to consistently use the subscripts for frame type and supperscripts for summation index. Do you think I should change it? } \jlk{On second reading, it is OK.}
to make sure that the current frame and foreground labels are aligned. By using different center spatial indices for background images and the current frame, we emulate camera jitter effect in the input.

\textbf{PTZ Camera Crop:}
Another challenging BGS scenario is PTZ camera operation. While such videos are very common in surveillance, 
they form only a small fraction of public datasets. Therefore, we introduce another data augmentation technique specific to this challenge.

Since PTZ videos do not have a static empty background frame, BSUV-Net \cite{tezcan2020bsuv} handles them differently than other categories. Instead of empty and recent backgrounds, the authors suggest to use recent and more recent background, where the recent background is computed as the median of 100 preceding frames and the more recent background is computed as the median of 30 such frames. To simulate this kind of behavior, we introduce two types of PTZ camera crops: (i) zooming camera crop, (ii) moving camera crop.

The zooming camera crop is defined as follows:
%
%\vspace{-1 ex}

{\small
\begin{equation*}
\begin{aligned}
    \anyfrcrop &= \cropfunc(\anyfr,\ i,\ j,\ \heightcrop,\ \widthcrop)
    \quad \text{ for all } 
    k \in \{\mathbf{\currentsub},\ \mathbf{\fgsub}\}, 
    \\
    \anyfrcrop &= \frac{1}{N^z} \sum_{n=0}^{N^z-1}{\mathbf{\widetilde{I}}_k^n}
    \quad \text{ for all } 
    k \in \{\mathbf{\emptysub},\ \mathbf{\recentsub}\}, \ \mathrm{where}
    \\
    {\mathbf{\widetilde{I}}_k^n} &= \resizefunc\Big(\cropfunc\Big(\anyfr,\ i,\ j,\ 
    \heightcrop(1 + nz_k),\  \widthcrop(1 + nz_k) \Big),\ \heightcrop,\ \widthcrop\Big) 
\end{aligned}
\end{equation*}
% %\vglue -1 ex
}%
\noindent where $z_\mathbf{\emptysub}, z_\mathbf{\recentsub}$ represent zoom factors for empty and recent backgrounds and $N^z$ represents the number of zoomed in/out frames to use in averaging.
In our experiments, we use  $-0.1 \! < \! z_\mathbf{\emptysub},\ z_\mathbf{\recentsub} \! < \! 0.1$ and $5 \! < \! N^z \! < \! 15$ to simulate real-world camera zooming.
$\resizefunc(\mathbf{I}, \heightcrop, \widthcrop)$ is an image resizing function that changes the resolution of $\mathbf{I}$ to $ (\widthcrop, \heightcrop)$ using bilinear interpolation. 
Note, that using positive values for $z_k$ simulates zooming in whereas using negative values simulates zooming out. Fig.~\ref{fig:aug}(d) shows an example of zoom-in.

Similarly, the moving camera crop is defined as follows:
%
%\vspace{-1 ex}
{\small
\begin{equation*}
\begin{aligned}
    \anyfrcrop &= \cropfunc(\anyfr,\ i,\ j,\ \heightcrop,\ \widthcrop)
    \quad \text{ for all } 
    k \in \{\mathbf{\currentsub},\ \mathbf{\fgsub}\},
    \\
    \anyfrcrop &= \frac{1}{N^m_\anysub} \sum_{n=0}^{N^m_\anysub-1}{\mathbf{\widetilde{I}}_k^n}
    \quad \text{ for all } 
    k \in \{\mathbf{\emptysub},\ \mathbf{\recentsub}\}, \ \mathrm{where}
    \\
    {\mathbf{\widetilde{I}}_k^n} &= \cropfunc(\anyfr,\ i+np,\ j+nq,\ \heightcrop,\ \widthcrop)
\end{aligned}
\end{equation*}
%\vglue -1 ex
}%
\noindent where $p, q$ are the vertical and horizontal shift amounts per frame and $N^m_\mathbf{\emptysub}, N^m_\mathbf{\recentsub}$ represent the number of empty and recent moving background crops to use for averaging. This simulates camera pan and tilt. 
%\piedit{ I understand pan, but how does it capture tilt?} 
In our experiments, we use  $-5<p,\ q<5$ and $5<N_\mathbf{E}^m, N_\mathbf{R}^m < 15$ to simulate real-world camera movements. 
%\jlk{Why do we need two separate $N^m$ here but not for $N^z$?}
% \ozan{Thats how I implemented them in the beginning. For zooming, I could have use single $z$ (instead of $z_k$) with $N^z_\mathbf{E}$ and $N^z_\mathbf{R}$} \jlk{So, in zooming we are using $N^z_\mathbf{R}=N^z_\mathbf{E}=N^z$, right?} \ozan{Yes. in Zoom in there is a single number used for averaging ($N^z$), put two different zoom-in ratios for empty and recent background ($z_\mathbf{E}, z_\mathbf{R}$)} \piedit{I am satisfied with Ozan's explanation.}

\subsection{Post-Crop Augmentations}
\label{postcrop}

In this section, we define several content-modifying augmentation techniques to compute $\emptyreffinal, \recentreffinal, \currentfrfinal, \fgfinal$ from $\emptyrefcrop, \recentrefcrop, \currentfrcrop, \fgcrop$. These augmentations can be applied after any one of the spatio-temporal crop augmentations.

\textbf{Illumination Difference:}
Illumination variations are common, especially in long videos, for example due to changes in natural light or lights being turned on/off. A temporal data augmentation technique to handle illumination changes was introduced in BSUV-Net \cite{tezcan2020bsuv} with the goal of increasing the network's generalization capacity for unseen videos. We use this augmentation here as well, formulated in our notation as follows:

%\vspace{-2 ex}
{\small
\begin{equation*}
    \anyfrfinal[i,\ j,\ c] = \anyfrcrop[i,\ j,\ c] + \mathbf{d}_k[c] 
    \text{ for }
    k \in \{\mathbf{\emptysub},\ \mathbf{\recentsub},\ \mathbf{\currentsub}\},
%    \text{ $\&$ }
    c = 1, 2, 3
\end{equation*}
%\vglue -1 ex
}%
\noindent where $\mathbf{d_E},\mathbf{d_R}, \mathbf{d_C} \in \mathbb{R}^3$ represent illumination offsets applied to RGB channels of the input images. 

\textbf{Intermittent-Object Addition:}
Another challenge for BGS are scenarios when objects enter a scene but then stop and remain static for a long time. Even very successful BGS algorithms, after some time, predict these objects as part of the background for they rely on recent frames to estimate the background model. BSUV-Net overcomes this challenge by using inputs from multiple time scales, however it still underperforms on videos with intermittently-static objects. To address this, we introduce another spatio-temporal data augmentation specific to this challenge.

\begin{figure}
  \includegraphics[width=\linewidth]{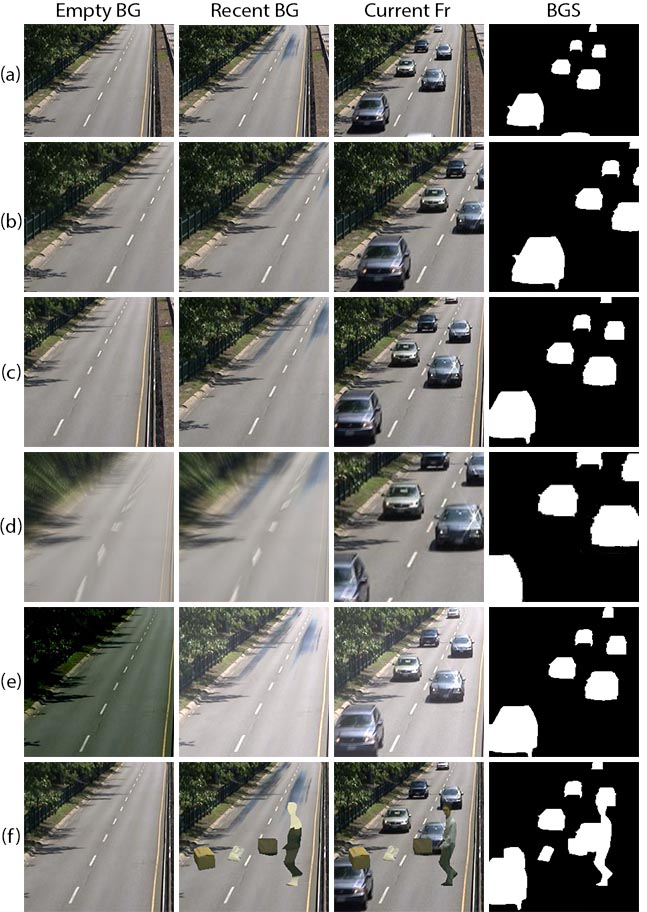}
  \caption{Image augmentation examples. Each row shows an example for one of the augmentations: (a) original input, (b) spatially-aligned crop, (c) randomly-shifted crop, (d) PTZ camera crop, (e) illumination difference, (f) intermittent-object addition.}
  \label{fig:aug}
%\vglue -3 ex
\end{figure}

We use a masking-based approach for intermittently-static objects as follows. In addition to the cropped inputs $\emptyrefcrop, \recentrefcrop, \currentfrcrop, \fgcrop$, we also use cropped inputs from videos with intermittently-static objects defined as $\emptyrefiom, \recentrefiom, \currentfriom \in \mathbb{R}^{\widthcrop \times \heightcrop \times 4}$ and $\fgiom \in \{0, 1\}^{\widthcrop \times \heightcrop}$. We copy foreground pixels from the intermittently-static input and paste them into the original input to synthetically create an intermittent object. This can be formulated as follows:
%
%\vspace{-1 ex}
{\small
\begin{equation*}
\begin{aligned}
    \emptyreffinal &= \emptyrefcrop\\
    \anyfrfinal &= \fgiom \odot \anyfriom + (1 - \fgiom) \odot \anyfrcrop
    \text{ for }
    k \in \{\mathbf{\currentsub}, \mathbf{\recentsub}\},\\
    \fgfinal &= \fgiom + (1-\fgiom) \odot \fgcrop
\end{aligned}
\end{equation*}
%\vglue -1 ex
}%
%
% \jlk{I think it would be more logical to write the last expression as follows: $\fgfinal = \fgiom + (1-\fgiom) \odot \fgcrop$}\ozan{Done!}
\noindent where $\odot$ denotes Hadamard (element-wise) product. Fig.~\ref{fig:aug}(f) shows an example of intermittent object addition. Note, that this augmentation requires prior knowledge of examples with intermittently-static objects which can be found in some public datasets.

\subsection{Combining Spatio-Temporal Augmentations}
\label{sec:aug_combine}

While the augmentations defined above can all be used by themselves to improve the BGS performance on related categories, combining multiple or even all of them might result in a better algorithm for a general unseen video of which the category is unknown. However, combining the crop algorithms is not trivial since it is not practical to apply more than one crop function to a single input. Thus, we use online augmentation, where we randomly augment every input while forming mini-batches. The augmentation steps are as follows:
\begin{enumerate}
    \item randomly select one of the spatial crop augmentations and apply it to the input,
    \item apply the illumination change augmentation using randomized illumination values,
    \item apply intermittent object addition to $p\%$ of the inputs.
\end{enumerate}
Clearly, a different combination of augmentations will be applied to the same input in different epochs. We hope this will significantly increase the generalization capacity of our network.

%-------------------------------------------------------------------------
\section{Video-Agnostic Evaluation Strategy for Supervised Algorithms}
\label{sec:crossval}

The most commonly used BGS datasets with a variety of scenarios and pixel-wise ground-truth annotations are CDNet-2014 \cite{goyette2012changedetection}, LASIESTA \cite{cuevas2016lasiesta} and SBMI2015 \cite{maddalena2015sbmi}. Among these 3 datasets, only CDNet-2014 has a well-maintained evaluation server, that keeps a cumulative performance record of the uploaded algorithms. Moreover, it has been the most widely-used dataset for BGS in recent years with publicly-available evaluation results for nearly all of the published BGS algorithms.

Since one of our aims is to compare the performance of \ourmodel{} with SOTA {\it video-agnostic} BGS algorithms on unseen videos, the availability of public results for these algorithms is critical. Therefore, we use CDNet-2014 as our evaluation dataset. CDNet-2014 is a comprehensive dataset that provides some ground-truth frames from all 53 videos to the public, but keeps others internally for algorithm comparison. Since it does not include any videos with no labeled frames, it is not directly suitable for testing of {\it video-agnostic} supervised algorithms. Consequently, most of the leading algorithms are either {\it video-optimized} or {\it video-group-optimized} and achieve near-perfect results by over-fitting the training data. However, these results are not generalizable to unseen videos \cite{tezcan2020bsuv, mandal2020scene, mandal20203dcd}. 
Several researchers addressed this problem by designing generalizable networks and evaluating their algorithms on unseen videos by using different videos in training and testing \cite{tezcan2020bsuv, mandal20193dfr, mandal2020scene, kim2020foreground, mandal20203dcd}. Yet, there is no common strategy for testing the performance of supervised BGS algorithms on CDNet-2014 for unseen videos. 
Some of the recent papers divide the dataset into two folds, train their algorithm on one of the folds and test on the other one. Since they only report the results on the test videos that they selected, their results might be biased towards their test set and not directly comparable with unsupervised algorithms. While BSUV-Net \cite{tezcan2020bsuv} proposes a {\it video-agnostic} evaluation strategy for the full CDNet-2014 dataset by using 18 training/testing video sets, this strategy might also be biased and is computationally expensive.
% \jlk{I simplified this especially that it is not clear what "task-specialized sets" means.}

In this paper, we introduce a simple and intuitive 4-fold cross-validation strategy for CDNet-2014.
We grouped all videos in the dataset and each category into 4 folds as evenly as possible (Table~\ref{table:traintest1}).
The proposed {\it video-agnostic} evaluation strategy is to train any supervised BGS algorithm on three of the folds and test on the remaining fold and replicate the same process for all 4 combinations. This approach will provide results on the full CDNet-2014 dataset which can be uploaded to the evaluation server to compare against SOTA. We believe this cross-validation strategy will be very beneficial for the evaluation of future BGS algorithms.

\begin{table}
\caption{Sets used in 4-fold cross-validation on CDNet-2014.
}
\smallskip
\smallskip
\centering
\scalebox{0.88}{
\begin{tabular}{|c|c|c|c|c|c|}
    \hline
     \textbf{category} & \textbf{video}  & $\mathbf{S_1}$& $\mathbf{S_2}$& $\mathbf{S_3}$& $\mathbf{S_4}$\\

    \hline \hline 
    \multirow{4}{*}{baseline} & highway & \cmark& & & \\ 
    \cline{2-6}
     & pedestrians &  & \cmark& & \\ 
    \cline{2-6}
     & office &  & &\cmark & \\ 
    \cline{2-6}
     & PETS2006 & & & &\cmark\\

    \hline \hline 
    \multirow{4}{*}{\makecell{bad\\weather}} & blizzard & \cmark& & &\\  
    \cline{2-6}
     & skating & &\cmark & &\\
     \cline{2-6}
     & wetSnow & & & \cmark&\\ 
    \cline{2-6}
     & snowFall & & & &\cmark\\

    \hline \hline 
    \multirow{6}{*}{\makecell{intermittent\\object\\motion}} & sofa &\cmark & & &\\ 
    \cline{2-6}
     & winterDriveway & &\cmark & &\\ 
    \cline{2-6}
     & parking & & & \cmark&\\ 
    \cline{2-6}
     & abandonedBox & & &\cmark &\\ 
    \cline{2-6}
     & streetLight & & & &\cmark\\ 
    \cline{2-6}
     & tramstop & & & &\cmark\\

    \hline \hline 
    \multirow{4}{*}{\makecell{low\\framerate}} & port 0.17fps & \cmark& & &\\ 
    \cline{2-6}
     & tramCrossroad 1fps & &\cmark & &\\ 
    \cline{2-6}
     & tunnelExit 0.35fps & & & \cmark&\\ 
    \cline{2-6}
     & turnpike 0.5fps & & & &\cmark\\

    \hline \hline 
    \multirow{4}{*}{PTZ} & continuousPan & \cmark& & &\\ 
    \cline{2-6}
     & intermittentPan & &\cmark & &\\ 
    \cline{2-6}
     & zoomInZoomOut & & & \cmark&\\ 
    \cline{2-6}
     & twoPositionPTZCam & & & &\cmark\\ 
     
    \hline \hline 
    \multirow{5}{*}{thermal} & corridor & \cmark& & & \\ 
    \cline{2-6}
     & lakeSide &\cmark & & & \\ 
    \cline{2-6}
     & library & &\cmark & &\\ 
    \cline{2-6}
     & diningRoom & & & \cmark&\\ 
    \cline{2-6}
     & park & & & &\cmark\\ 
    
    \hline \hline 
    \multirow{4}{*}{\makecell{camera\\jitter}} & badminton &\cmark & & &\\ 
    \cline{2-6}
     & traffic & & \cmark& &\\ 
    \cline{2-6}
     & boulevard & & &\cmark &\\ 
    \cline{2-6}
     & sidewalk & & & &\cmark\\ 
     
    \hline \hline 
    \multirow{6}{*}{shadow} & copyMachine &\cmark & & &\\
    \cline{2-6}
     & busStation & &\cmark & &\\  
    \cline{2-6}
     & cubicle & & & \cmark&\\ 
    \cline{2-6}
     & peopleInShade & & &\cmark &\\ 
    \cline{2-6}
     & bungalows & & & &\cmark\\ 
    \cline{2-6}
     & backdoor & & & &\cmark\\ 
     
    \hline \hline 
    \multirow{6}{*}{\makecell{dynamic\\background}}  & overpass &\cmark & & &\\ 
    \cline{2-6}
     & fountain02 &\cmark & & &\\ 
    \cline{2-6}
     & fountain01 & &\cmark & &\\ 
    \cline{2-6}
    & boats & &\cmark & &\\ 
    \cline{2-6}
     & canoe & & & \cmark&\\ 
    \cline{2-6}
     & fall & & & &\cmark\\

    \hline \hline 
    \multirow{6}{*}{\makecell{night\\videos}} & bridgeEntry & \cmark& & &\\ 
    \cline{2-6}
     & busyBoulvard &\cmark & & &\\ 
    \cline{2-6}
     & tramStation & &\cmark & &\\ 
    \cline{2-6}
     & winterStreet & &\cmark & &\\ 
    \cline{2-6}
     & fluidHighway & & & \cmark&\\ 
    \cline{2-6}
     & streetCornerAtNight & & & &\cmark\\

    \hline \hline 
    \multirow{4}{*}{turbulence} & turbulence0 & \cmark& & & \\ 
    \cline{2-6}
     & turbulence1 & &\cmark & &\\ 
    \cline{2-6}
     & turbulence2 & & &\cmark &\\ 
    \cline{2-6}
     & turbulence3 & & & & \cmark\\ 
     \hline
     
\end{tabular}}
%\vglue -3 ex
\label{table:traintest1}
\end{table}

% \jlk{The videos in the table are not listed alphabetically in each category, so how about arranging their order in a way that we have as "linear" check marks in each category as possible (with some duplicates). Otherwise reviewers may be tempted to say that we "cherry-picked".} \ozan{Thanks for pointing that. I organized the table}

%-------------------------------------------------------------------------
\vspace{4 ex}
\section{Experimental Results}
\label{sec:exp_res}

\subsection{Dataset and Evaluation Details}\label{sec:data_and_eval}

We evaluate the performance of our algorithm on CDNet-2014 \cite{goyette2012changedetection} using the evaluation strategy described in Section~\ref{sec:crossval}.
In CDNet-2014, the spatial resolution of videos varies from $320 \times 240$ to $720 \times 526$ pixels. The videos are labeled pixel-wise as follows: 1) foreground, 2) background, 3) hard shadow or 4) unknown motion. As suggested in \cite{goyette2012changedetection}, during evaluation we ignore pixels with unknown motion label and consider hard-shadow pixels as background.

In performance evaluation, we use metrics reported by CDNet-2014, namely recall ($Re$), specificity ($Sp$), false positive rate ($FPR$), false negative rate ($FNR$), percentage of wrong classifications ($PWC$), precision ($Pr$) and F-score ($F_1$). We also report two ranking-based metrics, ``average ranking'' ($R$) and ``average ranking across categories'' ($R_{cat}$), which combine all 7 metrics into ranking scores. A detailed description of these rankings can be found in \cite{goyette2012changedetection_old}. 
    
In order to better understand the performance of \ourmodel{} on unseen videos, we also performed a {\it cross-dataset} evaluation by training our model on CDNet-2014 and testing it on a completely different dataset, LASIESTA \cite{cuevas2016lasiesta}. LASIESTA is an extensive BGS dataset which includes 24 different videos from various indoor and outdoor scenarios. It includes a ``Simulated Motion'' category that is comprised of fixed-camera videos that are post-processed to mimic camera pan, tilt and jitter \cite{cuevas2016lasiesta}.
% \piedit{This reads well.}

\subsection{Training Details}

In order to train \ourmodel, we use similar parameters to the ones used for BSUV-Net. The same parameters are used for each of the four cross-validation folds. 
We used ADAM optimizer with a learning rate of $10^{-4}$, $\beta_1 = 0.9$, and $\beta_2=0.99$. The mini-batch size was 8 and the number of epochs was 200.
As the empty background frame, we used manually-selected frames introduced in \cite{tezcan2020bsuv}.  We used the median of preceding 100 frames as the recent background. 

In terms of spatio-temporal data augmentations, we use an online approach to randomly change the parameters under the following constraints. The random pixel shift between inputs is sampled from $\, \mathcal{U}(0, 5)$ where $\mathcal{U}(a, b)$ denotes uniform random variable between $a$ to $b$. The zooming-in ratios are sampled from $\mathcal{U}(0, 0.02)$ and $\mathcal{U}(0, 0.04)$ for recent and empty backgrounds, respectively, while the zooming-out ratios are sampled from $\mathcal{U}(-0.02, 0)$ and $\mathcal{U}(-0.04, 0)$.
We use $N^z = 10$. The horizontal pixel shift for moving camera augmentation is sampled from $\mathcal{U}(0, 5)$ with $N_\mathbf{\emptysub}^m=20$ and $N_\mathbf{\recentsub}^m=10$. We perform no vertical-shift augmentation since CDNet-2014 does not include any videos with vertical camera movement.
For illumination change, assuming $[0, 1]$ as the range of pixel values, we use $\mathbf{d_R}[k] = \mathbf{d_C}[k] = I + I_k$ where $I \sim \mathcal{N}(0, 0.1^2)$ and $I_k \sim \mathcal{N}(0, 0.04^2)$ for $k \in \{1, 2, 3\}$. Similarly, $\mathbf{d_E}[k] = \mathbf{d_C}[k] + I^E + I^E_k$ where $I ^E\sim \mathcal{N}(0, 0.1^2)$ and $I^E_k \sim \mathcal{N}(0, 0.04^2)$ for $k \in \{1, 2, 3\}$.
Lastly, for intermittent object addition, we always use the ``intermittent object motion'' inputs from the current training set and apply this augmentation to $p=10$\% of the inputs only. 
% Details of the training and evaluation implementation, with all of the defined augmentations, will be made publicly available upon publication. 
During inference, binary maps are obtained by thresholding network output at $\theta=0.5$.

\subsection{Ablation Study}

\begin{table*}[!htbp]
    \centering
    \caption{Comparison of  different spatio-temporal augmentations on CDNet-2014 based on F-score. SAC: spatialy-aligned crop, RSC: randomly-shifted crop, PTZ: PTZ camera crop, ID: illumination difference, IOA: intermittent object addition. The values in boldface font show the best performance for each category.
    }
    \smallskip
    \scalebox{0.8}{
        \begin{tabular}{ccccc|cccccccccccc}
            %\multicolumn{13}{c}{{F-score}} \\
            \!\!\makecell{SAC}\!\! & 
             \!\!\makecell{RSC}\!\!  & \!\!\makecell{PTZ}\!\! &\!\!\makecell{ID}\!\!&\!\!\makecell{IOA}\!\!& \!\!\! \makecell{Bad\\weather}\! & \!\! \makecell{Low\\framerate}\!\!\! & Night & \!\!\! PTZ & \!\!\!\!\!Thermal\!\! & \!\!\!\! Shadow\!\!&   \!\! \makecell{Int.\ obj.\\motion} &\!\!\!\! \makecell{Camera\\jitter} \! & \!\! \makecell{Dynamic\\backgr.}\!\!  & \makecell{Base-\\line} & \makecell{Turbu-\\lence} & Overall\\
             \hline
             \hline
              \cmark  &&  & & & 0.9442 & 0.7886 & 0.6982 & 0.6564 & 0.8960 & 0.9848 & 0.7732 & 0.8237 & 0.8517 & 0.9878 & 0.7285 & 0.8303\\
             \hline
             \cmark  &\cmark  & & & & 0.9439 & \textbf{0.8217} & 0.6941 & 0.6304 & 0.8854 & 0.9818 & 0.7620 & 0.9043 & 0.8745 & 0.9865 & 0.7354 & 0.8382\\
             \hline
             \cmark  &&  \cmark & & & 0.9315 & 0.7961 & 0.6557 & \textbf{0.6815} & 0.8905 & 0.9795 & 0.7458 & 0.8999 & 0.8674 & 0.9838 & 0.7409 & 0.8339\\
              \hline
             \cmark  &&  &\cmark  & & \textbf{0.9489} & 0.7606 & \textbf{0.7605} & 0.6579 & \textbf{0.9024} & \textbf{0.9855} & 0.7503 & 0.8270 & 0.8364 & 0.9874 & 0.7341 & 0.8319\\
              \hline
             \cmark  &&  &   & \cmark & 0.9456 & 0.7550 & 0.7233 & 0.6383 & 0.8997 & 0.9836 & \textbf{0.9312} & 0.8359 & 0.8709 & \textbf{0.9883} & 0.7023 & 0.8431\\
             \hline
             \cmark  &\cmark  &\cmark &\cmark &\cmark & 0.9272 & 0.8114 & 0.6841 & 0.6725 & 0.8960 & 0.9811 & 0.8489 & \textbf{0.9163} & \textbf{0.8848} & 0.9834 & \textbf{0.8056} & \textbf{0.8556}\\
             \hline
        \end{tabular}
        }
        \label{table:ablation}
\end{table*}

\begin{figure*}[!htbp]
  \includegraphics[width=0.95\linewidth]{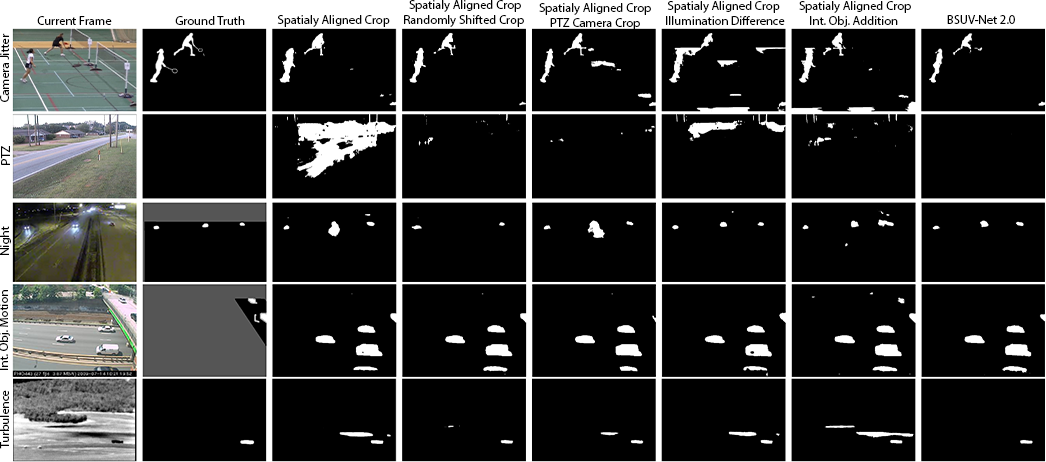}
  \centering
  \caption{Visual comparison of  different spatio-temporal augmentations on sample frames. Columns 1 and 2 show the current frame and its ground truth. The subsequent columns show predictions of our network trained with augmentations listed at the top of each column. The last column shows predictions of \ourmodel{} trained using a combination of all proposed augmentations. The dark-gray areas of ground-truth images represent pixels outside of CDNet-2014 regions of interest.}
  \label{fig:ablation}
%\vglue -3 ex
\end{figure*}

Let's assess the impact of each spatio-temporal data augmentation method defined in Section~\ref{sec:augmentations}. As the baseline network, we use BSUV-Net with only spatially-aligned crop augmentation and random Gaussian noise sampled from $\mathcal{N}(0, 0.01^2)$.
We evaluate the proposed spatio-temporal augmentations against this baseline by including the spatially-aligned crop among spatial crop augmentations, as explained in Section~\ref{sec:aug_combine}.
% \jlk{Do we use random Guassian noise? My sentence suggests this but the explanation in Section IV-D does not.}
% \ozan{Yes, we do. Since Section IV-D focuses on spatio-temporal augmentations and random Gaussian noise is a standard procedure applied in almost all deep learning models, I didn't include it there.}
%\jlk{Did you mean spatial crop? Can PTZ crop be combined with spatially-aligned crop? From Section IV-D I understood that not.}
%\ozan{Combination of augmentations is done as explained in steps in Section~\ref{sec:aug_combine}. We are not applying them to the same input, we are randomly selecting one of the augmentations for each input of the network.}
In PTZ camera crop, for each input, we randomly select one of the following: zooming in, zooming out, moving right or moving left. 
% A combination of data augmentations is performed as described in Section~\ref{sec:aug_combine} with the addition of random Gaussian noise applied after the last step.
Table~\ref{table:ablation} shows the category-wise F-score results for CDNet-2014.
All results have been calculated locally (not on CDNet-2014 evaluation server) for those CDNet-2014 frames with publicly-available ground truth and we report the median of results for every $5^{th}$ epoch between $150^{th}$ and $200^{th}$ epochs to disregard small fluctuations in the learning process.
Fig.~\ref{fig:ablation} shows some visual results for these algorithms for 5 videos.
It can be observed that each augmentation type significantly improves the performance on related categories (randomly shifted crop -- on ``Camera jitter'', PTZ camera crop -- on ``PTZ'', illumination difference -- on ``Shadow'', intermittent object addition -- on ``Intermittent object motion''), but combining all augmentations decreases the performance significantly on some categories (e.g., night and intermittent object motion). We believe this is due to trade-offs between the effects of different augmentations. For example, when a static background object starts moving it should be labeled as foreground, but a network trained with a randomly-shifted crop augmentation can confuse this input with an input from the ``Camera jitter'' category and continue labeling the object as background. Still, the overall performance (last column in Table~\ref{table:ablation}) of \ourmodel{} that uses all augmentations handily outperforms the overall performance for individual augmentations.

Since BGS is often applied as a pre-processing step in real-time video processing applications, computation speed is critical. As discussed in our previous work \cite{tezcan2020bsuv}, one of the main bottlenecks of BSUV-Net is the computation of FPM for each channel -- it decreases the overall computation speed significantly. On the other hand, either removing the FPM channel or predicting BGS by thresholding the FPM channel alone decreases the performance to values that are lower than that of some unsupervised algorithms \cite{tezcan2020bsuv}. In this work, we show that the performance of our model, even without the FPM channel but with augmentations, is better than the current state-of-the-art. We call this version of \ourmodel{}, which uses 9 instead of 12 channels on input, \ourmodelfast{}. Table~\ref{table:speed} shows a speed and performance comparison of the two versions. Clearly, while \ourmodelfast{} has slightly lower performance, it can be used in real-time applications at $320 \times 240$ spatial resolution, which is very similar to the resolution used in training. For higher-resolution videos, one can easily feed decimated frames into \ourmodelfast{} and interpolate the resulting BGS predictions to the original resolution.

\begin{table}[!htbp]
    % \vglue -1 ex
    \centering
    \caption{
    Efficiency vs performance trade-off for \ourmodel{} on CDNet-2014. FPS is calculated using PyTorch 1.3 implementation on a node with single Nvidia Tesla P100 GPU.
    }
    \smallskip
    \scalebox{0.85}{
    \begin{tabular}{c|ccccc}
        %\multicolumn{13}{c}{{F-score}} \\
        \hline
        & \multirow{2}{*}{$Re$}  & \multirow{2}{*}{$Pr$} & \multirow{2}{*}{$F_1$} & \multicolumn{2}{c}{$FPS$}\\
        & & & &\!\!\! $320 \times 240$ \!\!\!&\!\!\! $640 \times 480$\!\!\! \\
         \hline
         \ourmodel & 0.85 & 0.89 & 0.86 & $\sim 6$ & $\sim 2.5$ \\
         \ourmodelfast & 0.84 & 0.84 & 0.81 & $\sim 29$ & $\sim 13$\\
         \hline
    \end{tabular}
    }
    %\vglue -2 ex
    \label{table:speed}
\end{table}

\subsection{Comparison with State of the Art}
%%%%%%%%%%%%%%%%%%%%%%%%%%%%%%%%%%%%%%%%%%%%%%%%%%%%%%%%%%%%%%
\begin{table*}[!htbp]
  \centering
  \caption{
  Official comparison of top BGS algorithms evaluated on {\bf unseen videos} from CDNet-2014.
  }
  \smallskip
  \scalebox{0.95}{
        \begin{tabular}{c|cc|ccccccc}
            \hline
             Method & $R$ & $R_{cat}$ & $Re$ & $Sp$ & $FPR$ & $FNR$ & $PWC$ & $Pr$ & $F_1$\\
             \hline
             {\bf BSUV-Net 2.0} & \textbf{7.71} & \textbf{7.73} & 0.8136 & \textbf{0.9979} & \textbf{0.0021} & 0.1864 & \textbf{0.7614} & \textbf{0.9011} & \textbf{0.8387}\\
             {\bf Fast BSUV-Net 2.0} & 	7.86 & 10.36 & {0.8181} & 0.9956 & 0.0044 & {0.1819} & 0.9054 & 0.8425 & 0.8039\\
             { BSUV-Net} + SemanticBGS \cite{tezcan2020bsuv} & {9.57} & 14.27 & {0.8179} & 0.9944 & 0.0056 & {0.1821} & 1.1326 & {0.8319} & {0.7986}\\
             IUTIS-5 + SemanticBGS \cite{braham2017semanticbgs} & 9.71 & 11.91 & 0.7890 &	{0.9961} &	{0.0039} &	0.2110 &	{1.0722} &	0.8305 & 0.7892\\
             IUTIS-5 \cite{bianco2017iutis}& 12.14 & {10.91} & 0.7849 &	0.9948 &	0.0052 &	0.2151 &	1.1986 &	0.8087 & 0.7717	\\
             %\hline
             { BSUV-Net} \cite{tezcan2020bsuv}  & {9.71} & {14.00} & \textbf{0.8203} &	0.9946 &	0.0054 &	\textbf{0.1797} &	{1.1402} &	{0.8113} &	{0.7868}\\
            %  SWCD \cite{icsik2018swcd}& 16.43 & 20.00 & 0.7839 &	0.9930 &	0.0070 &	0.2161 &	1.3414 &	0.7527 &	0.7583\\
             WisenetMD \cite{lee2018wisenetmd}& 17.29 & 15.82 & 0.8179 &	0.9904 &	0.0096 &	0.1821 &	1.6136 &	0.7535 &	0.7668\\ 
            %  PAWCS \cite{st2015pawcs}& 14.71 & 16.09 & 0.7718 &	{0.9949} &	{0.0051} &	0.2282 &	1.1992 &	0.7857 &	0.7403\\
             FgSegNet v2 \cite{lim2018fgsegnet}&	45.57 &45.09&	0.5119&	0.9411&	0.0589	&0.4881&	7.3507&	0.4859&	0.3715\\
             \hline
        \end{tabular}
  }
  %\vglue -1 ex
  \label{table:overall_results}
\end{table*}
\begin{table*}[!htbp]
  \centering
  \caption{Official comparison of top BGS algorithms according to the per-category F-score on {\bf unseen videos} from CDNet-2014.
  }
  \smallskip
  \scalebox{0.8}{
        \begin{tabular}{c|ccccccccccc|c}
            \hline
             Method & \!\!\! \makecell{Bad\\weather}\! & \!\! \makecell{Low\\framerate}\!\!\! & Night & \!\!\! PTZ & \!\!\!\!\!Thermal\!\! & \!\!\!\! Shadow\!\!&   \!\! \makecell{Int.\ obj.\\motion} &\!\!\!\! \makecell{Camera\\jitter} \! & \!\! \makecell{Dynamic\\backgr.}\!\!  & \makecell{Base-\\line} & \makecell{Turbu-\\lence} & Overall\\
             \hline
             \!\!{\bf BSUV-Net 2.0} & 0.8844 & {\bf0.7902} & 0.5857 & {\bf 0.7037} & {\bf 0.8932} & {0.9562} & 0.8263 & {\bf 0.9004} & 0.9057 & 0.9620 & 0.8174 & {\bf 0.8387}\\
             \!\!{\bf Fast BSUV-Net 2.0} & {\bf0.8909} & 0.7824 & 0.6551 & 0.5014 & 0.8379 & 0.8890 & {\bf 0.9016} & 0.8828 & 0.7320 & {\bf 0.9694} & 0.7998 & 0.8039\\
             \!\!{ BSUV-Net} + SemanticBGS & {0.8730} & 0.6788 & {0.6815} & {0.6562} & {0.8455} & {\bf 0.9664} & 0.7601 & 0.7788 & 0.8176 & {0.9640} & 0.7631 & {0.7986}\\
             IUTIS-5 + SemanticBGS & 0.8260 & 0.7888 & 0.5014 & 0.5673 & 0.8219 & 0.9478 & {0.7878} & {0.8388} & {\bf 0.9489} & 0.9604 & 0.6921 & 0.7892\\
             IUTIS-5 & 0.8248 & {0.7743} & 0.5290 & 0.4282 & 0.8303 & 0.9084 & 0.7296 
             & 0.8332 & 0.8902 & 0.9567 & {0.7836} & 0.7717\\
             { BSUV-Net}  & {0.8713}&	0.6797&	{\bf 0.6987}&	{0.6282}&	{0.8581}&	0.9233&	0.7499&	0.7743&	0.7967&	{0.9693}&	0.7051	& 0.7868\\
             RTSS & 0.8662 & 0.6771 & 0.5295 & 0.5489 & 0.8510 & {0.9551} & {0.7864} & {0.8396} & {0.9325} & 0.9597 & 0.7630 & {0.7917}\\
             WisenetMD & 0.8616 & 0.6404 & 0.5701 & 0.3367 &  0.8152 & 0.8984 & 0.7264 
             & 0.8228  & 0.8376 & 0.9487 & {\bf 0.8304}& 0.7535\\
             FgSegNet v2 & 0.3277 & 0.2482 & 0.2800 & 0.3503 & 0.6038 & 0.5295 & 0.2002 & 0.4266 & 0.3634 & 0.6926 & 0.0643 & 0.3715\\
             \hline
        \end{tabular}
  }
  \label{table:cat_results}
  %\vglue -2 ex
\end{table*}

\begin{table}[!htbp]
  \centering
  \caption{F-score comparison of \ourmodel{} with {\it video-agnostic} supervised BGS algorithms that are not reported in \href{changedetection.net}{\tt changedetection.net.} Each column shows test performance of the algorithm by using the training/testing split provided in respective paper. %\jlk{We should explain that columns mean different splits.}\ozan{Is it better?} \piedit{I added the word "Training/Testing Split" to make it more clear.}
  }
  \smallskip
  \scalebox{0.9}{
        \begin{tabular}{c|cccc}
            \multicolumn{1}{c}{} & \multicolumn{4}{c}{Training/Testing Split} \\
            \hline
            Method & 3DCD & 3DFR & ChangeDet  & Kim \\
             \hline
             \!\!{\bf BSUV-Net 2.0} & {\bf 0.89} & {\bf 0.89} & {\bf 0.90} & {\bf 0.88}\\
             \!\!{\bf Fast BSUV-Net 2.0} & 0.84 & 0.84 & 0.86 & 0.88\\
             3DCD \cite{mandal20203dcd} & 0.86 & - & - & -  \\
             3DFR \cite{mandal20193dfr} & - & 0.86 & - & -  \\
             ChangeDet \cite{mandal2020scene}& - & - & 0.84 & -\\
             Kim \cite{kim2020foreground} & - & - & - & 0.86\\
             \hline
        \end{tabular}
  }
  \label{table:eval_others}
  %\vglue -3ex
\end{table}
Table~\ref{table:overall_results} shows the performance of \ourmodel{} and \ourmodelfast{} compared to state-of-the-art BGS algorithms that are designed for and tested on unseen videos. We did not include the results of {\it video-} or {\it video-group-optimized} algorithms since it is not fair to compare them against {\it video-agnostic} algorithms. This table shows official results computed by CDNet-2014 evaluation server\footnote{Full results can be accessed from \href{http://jacarini.dinf.usherbrooke.ca/results2014/870/}{\tt jacarini.dinf.}
\href{http://jacarini.dinf.usherbrooke.ca/results2014/870/}{\tt usherbrooke.ca/results2014/870/} and
\href{http://jacarini.dinf.usherbrooke.ca/results2014/871/}{\tt jacarini.dinf.}
\href{http://jacarini.dinf.usherbrooke.ca/results2014/871/}{\tt usherbrooke.ca/results2014/871/}
 }, so the results of our models slightly differ from those in 
 Tables~\ref{table:ablation} and \ref{table:speed} (different ground-truth frames). 
%  \piedit{Just to confirm: are you referring to the value 0.8556 for Overall with all augmentations in Table 2 as compared to the value 0.8387 for F1 score of BSUV-Net 2.0 in Table 4?}\ozan{Yes}
 We compare \ourmodel{} with some of the top-performing {\it video-agnostic algorithms} reported by this server. RTSS \cite{zeng2019rtss}, 3DCD \cite{mandal20203dcd}, 3DFR \cite{mandal20193dfr}, ChangeDet \cite{mandal2020scene} and Kim {\it et al.} \cite{kim2020foreground} are not included in this table since their results are not reported.
 {\it Video-agnostic} results of FgSegNet v2 are taken from \cite{tezcan2020bsuv}.
\ourmodel{} outperforms all SOTA algorithms by at least $\sim$5\% in terms of F-score (0.8387 versus 0.7986 in Tables~\ref{table:overall_results}, \ref{table:cat_results}). \ourmodelfast{} also outperforms all state-of-the-art algorithms while being $\sim$5 times faster than \ourmodel{} during inference (Table~\ref{table:speed}). Table~\ref{table:cat_results} shows the comparison of $F_1$ results for each category. This table includes RTSS using results reported in the paper \cite{zeng2019rtss}. In 7 out of 11 categories, either \ourmodel{} or \ourmodelfast{} achieve the best performance, including most of the categories that we designed the augmentations for (an exception is the ``Night'' category). However, note that the best-performing algorithm in the ``Night'' category is BSUV-Net which uses only the illumination-difference augmentation. Thus, it focuses on videos with illumination differences such as night videos. 

\begin{figure*}[!htbp]
  \includegraphics[width=0.95\linewidth]{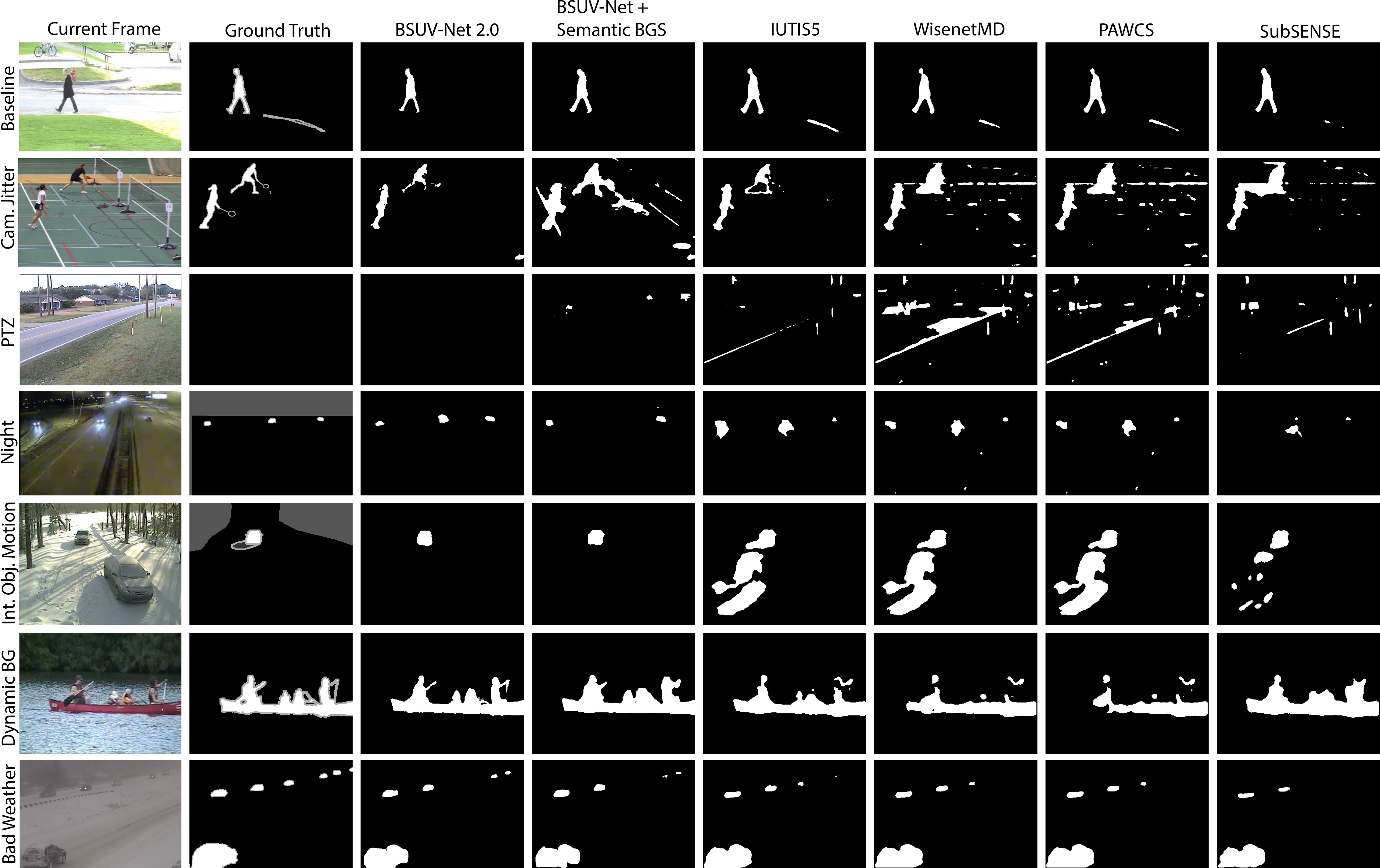}
  \centering
  \caption{Qualitative comparison of top BGS algorithms on sample frames from different categories of CDNet-2014.}
  \label{fig:sota}
%\vglue -3 ex
\end{figure*}

Fig.~\ref{fig:sota} qualitatively compares the performance of \ourmodel{} with state-of-the-art video-agnostic BGS algorithms on a few example videos from CDNet-2014. \ourmodel{} clearly produces the best visual results in a variety of scenarios. Results for \textit{Camera jitter} and \textit{PTZ} categories show the effectiveness of \ourmodel{} in removing false positives resulting from camera motion. In the example from \textit{Intermittent object motion} category, the car on the left is starting to back-up from the driveway and most of the algorithms produce false positives at the location where the car was parked whereas \ourmodel{} successfully eliminates these false positives. Results for \textit{Dynamic background} show that \ourmodel{} is very effective in accurately delineating the boundary between foreground objects and the background.

As discussed in Section~\ref{sec:related}, 3DCD \cite{mandal20203dcd}, 3DFR \cite{mandal20193dfr}, ChangeDet \cite{mandal2020scene} and Kim {\it et al.} \cite{kim2020foreground} are also among the best {\it video-agnostic} supervised algorithms, however each reports performance on a different subset of CDNet-2014, with the algorithm trained on the remaining videos. Table~\ref{table:eval_others} shows the comparison of \ourmodel{} with these algorithms using the training/testing splits provided in respective papers in each column. \ourmodel{} clearly outperforms all four competitors, while \ourmodelfast{} beats 2 out of 4, and does so with real-time performance.

% \vspace{3 ex}
\subsection{Cross-Dataset Evaluation}
\label{sec:crossdataset}

% \begin{table*}[htbp]
%   \centering
%   \caption{Comparison of the cross-dataset performance of \ourmodel{} with the top performing BGS algorithms on LASIESTA
%   }
%   \smallskip
%   \scalebox{0.9}{
%         \begin{tabular}{c|cccccccccc|c}
%             \hline
%              Method & I{\textunderscore}SI & I{\textunderscore}CA & I{\textunderscore}OC & I{\textunderscore}IL & I{\textunderscore}MB & I{\textunderscore}BS & O{\textunderscore}CL & O{\textunderscore}RA & O{\textunderscore}SN  & O{\textunderscore}SU & Overall\\
%              \hline
%             {\bf BSUV-Net 2.0} & 0.9151 & 0.6785& 0.9565& 0.8817& 0.8075& 0.7681& 0.9281& 0.9449& 0.8423& 0.7911& 0.8514\\
%             %  {\bf Fast BSUV-Net 2.0} & & & & & & & & & & & \\
%              Cuevas & 0.8805 &0.8444 &0.7806 &0.6487 &0.9373 &0.6644 &0.9276 &0.8669 &0.7786 & 0.7221&0.8051 \\
%              Haines &0.8876 &0.8938 &0.9223 &0.8491 &0.8440 &0.6809 &0.8267 &0.8907 &0.1740 &0.8568 &0.7826 \\
%              Maddalena &0.9484 &0.8573 &0.9540 &0.2105 &0.9122 &0.4018 &0.9708 &0.8972 &0.8104 &0.8792 &0.7842 \\
%              \hline
%         \end{tabular}
%   }
%   \label{table:lasiesta}
% \end{table*}

\begin{table*}[htbp]
  \centering
  \caption{Per-category F-score comparison of the cross-dataset performance of \ourmodel{} with the top-performing unsupervised BGS algorithms on LASIESTA}
  \smallskip
  \scalebox{1.0}{
        \begin{tabular}{c|cccccccccc|c}
            \hline
             Method & ISI & ICA & IOC & IIL & IMB & IBS & OCL & ORA & OSN  & OSU & Overall\\
             \hline
            {\bf BSUV-Net 2.0} & 0.92 & 0.68& \textbf{0.96}& \textbf{0.88}& 0.81& \textbf{0.77}& \textbf{0.93}& \textbf{0.94}& \textbf{0.84}& 0.79& \textbf{0.85}\\
            %  {\bf Fast BSUV-Net 2.0} & & & & & & & & & & & \\
             Cuevas \cite{berjon2018cuevas} & 0.88 &0.84 &0.78 &0.65 &\textbf{0.93} &0.66 &\textbf{0.93} &0.87 &0.78 & 0.72&0.81 \\
             Haines \cite{haines2013background} &0.89 &\textbf{0.89} &0.92 &0.85 &0.84 &0.68 &0.83 &0.89 &0.17 &0.86 &0.78 \\
             Maddalena2 \cite{maddalena2012sobs} &\textbf{0.95} &0.86 &0.95 &0.21 &0.91 &0.40 &0.97 &0.90 &0.81 &\textbf{0.88} &0.78 \\
              Maddalena1 \cite{maddalena2008self} &0.87 &0.85 &0.91 &0.61 &0.76 &0.42 &0.88 &0.84 &0.58 &0.80 & 0.75 \\
             \hline
        \end{tabular}
  }
  \label{table:lasiesta}
\end{table*}

\begin{table*}[htbp]
  \centering
  \caption{Per-video F-score comparison of \ourmodel{} trained on two different configurations with the top-performing supervised BGS algorithms on unseen videos from LASIESTA. \ourmodel{} is trained only on CDNet-2014, as explained in Sec.~\ref{sec:crossdataset}. \ourmodel{}$^*$ is trained on videos from LASIESTA that are not from the test set as described in \cite{mandal20203dcd}.
  }
  \smallskip
  \scalebox{0.95}{
        \begin{tabular}{c|cccccccccc|c}
            \hline
             Method & ISI-2 & ICA-2 & IOC-2 & IIL-2 & IMB-2 & IBS-2 & OCL-2 & ORA-2 & OSN-2  & OSU-2 & Overall\\
             \hline
            {\bf BSUV-Net 2.0}  &0.89 &0.60 &0.95 &0.89 &0.76 &0.69 &0.89 &0.93 &\textbf{0.70} &0.91 &0.82 \\
            {\bf BSUV-Net 2.0$^*$}  &\textbf{0.98} &\textbf{0.99} &\textbf{0.97} &\textbf{0.97} &\textbf{0.88} &\textbf{0.95} &\textbf{0.97} &\textbf{0.97} &0.55 &\textbf{0.95} &\textbf{0.92} \\
            % {\bf BSUV-Net 2.0$^{**}$}  & \textbf{0.98} &0.98 &\textbf{0.97} &0.96 &0.80 &\textbf{0.95} &\textbf{0.97} &\textbf{0.98} &0.53 &\textbf{0.95} & 0.91 \\
            %  {\bf Fast BSUV-Net 2.0} & & & & & & & & & & & \\
             3DCD \cite{mandal20203dcd}  &0.86 &0.49 &0.93 &0.85 &0.79 &0.87 &0.87 &0.87 &0.49 &0.83 &0.79  \\
             FgSegNet v2 \cite{lim2018fgsegnetv3} &0.53 &0.58 &0.25 &0.41 &0.63 &0.25 &0.54 &0.54 &0.05 &0.29 & 0.41  \\
             FgSegNet-M \cite{lim2018fgsegnet} &0.56 &0.55 &0.65 &0.42 &0.56 &0.19 &0.28 &0.18 &0.01 &0.33 & 0.37 \\
            %  Cuevas2  &0.84 &0.78 &0.86 &0.65 &0.92 &0.62 &0.90 &0.89 &0.63 &0.77 & 0.79 \\
            %  Haines  &0.81 &0.87 &\textbf{0.95} &0.81 &0.71 &0.73 &0.96 &0.96 &0.04 &0.90 & 0.77 \\
            %  Maddalena2  &\textbf{0.94} &\textbf{0.87} &\textbf{0.95} &0.23 &0.85 &0.40 &0.98 &0.96 &0.71 &0.88 & 0.78 \\
            %  Maddalena1  &0.85 &0.74 &0.85 &0.37 &0.68 &0.45 &0.85 &0.86 &0.46 &0.86 & 0.70 \\
             \hline
        \end{tabular}
  }
  \label{table:lasiesta_si}
\end{table*}

\begin{table}[htbp]
  \centering
  \caption{F-score comparison of cross-dataset performance of different spatio-temporal augmentations on \textit{Moving camera} and \textit{Simulated motion} videos of LASIESTA. SAC: spatialy-aligned crop, RSC: randomly-shifted crop, PTZ: PTZ camera crop.}
  \smallskip
  \scalebox{0.88}{
        \begin{tabular}{ccc|cccc}
            \hline
             \makecell{SAC} \!\! & 
             \!\! \makecell{RSC} \!\! & \!\! \makecell{PTZ} & \makecell{Indoor\\pan \& tilt} &  \makecell{Outdoor\\pan \& tilt} & \makecell{Indoor\\jitter} &  \makecell{Outdoor\\jitter}\\
             \hline
             \cmark & & & 0.48 & 0.56 & 0.81 & 0.75\\
             \cmark & \cmark & & 0.52 & 0.42 & \textbf{0.88} & \textbf{0.85}\\
             \cmark & & \cmark & \textbf{0.58} & \textbf{0.58} & 0.84 & 0.50\\
             \hline
        \end{tabular}
    }
  \label{table:lasiesta_moving}
\end{table}

In this section, we perform a cross-dataset evaluation to show the generalization capacity of \ourmodel{}. We train \ourmodel{} using CDNet-2014 videos from  $\bf{S_2}$, $\bf{S_3}$, $\bf{S_4}$ sets shown in Table~\ref{table:traintest1} and use $\bf{S_1}$ as a validation set to select the best performing epoch. Then, we evaluate the results on a completely different dataset, LASIESTA \cite{cuevas2016lasiesta}\footnote{The empty backgrounds of LASIESTA videos are computed automatically as the median of all frames in the video. The recent backgrounds are computed similarly to CDNet-2014, as the median of previous 100 frames.}. Table~\ref{table:lasiesta} shows the comparison of \ourmodel{} with top-performing unsupervised algorithms reported in \cite{cuevas2016lasiesta}. Since the authors reported results only for categories of LASIESTA recorded with static cameras, we report results only on these categories. Clearly, \ourmodel{} outperforms its competitors on a completely unseen dataset by a significant margin.

In \cite{mandal20203dcd}, Mandal {\it et al.}  performed a \textit{video-agnostic} evaluation of some supervised learning algorithms by training with 10 of the LASIESTA videos and evaluating on 10 unseen videos from LASIESTA. Table~\ref{table:lasiesta_si} shows a comparison of \ourmodel{} with unseen video performance of the algorithms reported in \cite{mandal20203dcd}. We show the results of \ourmodel{} trained with two different datasets. \ourmodel{} row shows the results of cross-dataset training whereas \ourmodel{}$^*$ row shows the results of using the same training set that is used in \cite{mandal20203dcd} for a fair comparison.
\ourmodel{} achieves significantly better results than state of the art even if the training set does not include any videos from LASIESTA. Since we train \ourmodel{}$^*$ with videos from LASIESTA, it performs even better than \ourmodel{}. This shows that the proposed spatio-temporal data augmentations are not specific to CDNet-2014 and can be very effective on other datasets as well.
Note that the performance of \ourmodel{} is significantly better than that of \ourmodel{}$^*$ on OSN-2, an outdoor video recorded in heavy snow. This is due to the fact that the training videos of LASIESTA do not include a heavy-snow video, however the training set of CDNet-2014 does. This also shows the importance of scene variety in the training dataset. Both Table~\ref{table:lasiesta} and \ref{table:lasiesta_si} clearly show that \ourmodel{} is not specific to a dataset that it was trained on, but can successfully predict BGS of an unseen video.

In addition to the videos reported in Table~\ref{table:lasiesta}, LASIESTA includes several BGS videos that are either recorded with a moving camera or post-proccessed to look like they were recorded with a moving camera. We group these videos under 4 categories:
\begin{enumerate}
  \item \textit{Indoor pan \& tilt videos} (IMC-1, ISM-1, ISM-2, ISM-3),
  \item \textit{Outdoor pan \& tilt videos} (OMC-1, OSM-1, OSM-2, OSM-3),
  \item \textit{Indoor jitter videos} (IMC-2, ISM-4, ..., ISM-12),
  \item \textit{Outdoor jitter videos} (OMC-2, OSM-4, ..., OSM-12).
\end{enumerate}
Table~\ref{table:lasiesta_moving} shows the F-score comparison of \ourmodel{} trained with different combinations of spatio-temporal data augmentations on these 4 categories. As expected, the randomly shifted crop augmentation achieves the best performance for videos with camera jitter whereas the PTZ augmentation achieves the best results for PTZ category. This further shows that the impact of spatio-temporal data augmentations is generalizable to different datasets.

%\vspace{-1 ex}
\section{Conclusion}

While background subtraction algorithms achieve remarkable performance today, they still often fail in challenging scenarios such as shaking or panning/tilting/zooming cameras, or when moving objects stop for an extended period of time. In the case of supervised algorithms, this is largely due to the limited availability of labeled videos recorded in such scenarios -- it is difficult to train end-to-end deep-learning algorithms for unseen videos. To address this, we introduced several spatio-temporal data augmentation methods to synthetically increase the number of inputs in such scenarios. Specifically, we introduced new augmentations for PTZ, camera jitter and intermittent object motion scenarios, and achieved significant performance improvements in these categories and, consequently, a better overall performance on CDNet-2014 dataset. We also introduced a real-time version of \ourmodel{} which still performs better than state-of-the-art methods and we proposed a 4-fold cross-validation data split for CDNet-2014 for easier comparison of future algorithms. Finally, we demonstrated a strong generalization capacity of \ourmodel{} using cross-dataset evaluation on LASIESTA in which the proposed model significantly outperforms the current state-of-the-art methods on a completely unseen dataset.

% \clearpage
{\small
\bibliographystyle{ieee_fullname}
\bibliography{strings, ref}
}

\end{document}